\documentclass[conference]{IEEEtran}
\IEEEoverridecommandlockouts
\usepackage{cite}
\usepackage{amsmath,amssymb,amsfonts}
\usepackage{algorithmic}
\usepackage{graphicx}
\usepackage{textcomp}
\usepackage{xcolor}
\def\BibTeX{{\rm B\kern-.05em{\sc i\kern-.025em b}\kern-.08em
    T\kern-.1667em\lower.7ex\hbox{E}\kern-.125emX}}

\usepackage{siunitx}
\usepackage{makecell}

\usepackage{booktabs}
\usepackage{hyperref}
\usepackage{mathtools}

\newcommand{\chg}[1]{{#1}}

\begin{document}
\bstctlcite{IEEEexample:BSTcontrol}

\title{Automated Static Camera Calibration with Intelligent Vehicles
\thanks{Part of this work was financially supported by the Federal Ministry for Economic Affairs and Climate Action of Germany within the program "New Vehicle and System Technologies" (project LUKAS, grant number 19A20004F).}
}

\author{\IEEEauthorblockN{Alexander Tsaregorodtsev\IEEEauthorrefmark{1}, Adrian Holzbock\IEEEauthorrefmark{1}, Jan Strohbeck\IEEEauthorrefmark{1}, Michael Buchholz\IEEEauthorrefmark{1}, and Vasileios Belagiannis\IEEEauthorrefmark{2}}
\IEEEauthorblockA{\IEEEauthorrefmark{1}\textit{Institute for Measurement-, Control-, and Microtechnology} \\
\textit{Ulm University}, Germany\\
%Ulm, Germany \\
{\tt \{firstname\}}.{\tt \{lastname\}}@uni-ulm.de}
%\and
%\IEEEauthorblockN{Vasileios Belagiannis}
\IEEEauthorblockA{\IEEEauthorrefmark{2}\textit{Chair of Multimedia Communications and Signal Processing} \\
\centerline{\textit{Friedrich-Alexander-Universität Erlangen-Nürnberg}, Germany}\\
%Erlangen, Germany \\
vasileios.belagiannis@fau.de}
}

\maketitle
\thispagestyle{empty}
\pagestyle{empty}

%%%%%%%%%%%%%%%%%%%%%%%%%%%%%%%%%%%%%%%%%%%%%%%%%%%%%%%%%%%%%%%%%%%%%%%%%%%%%%%%
\begin{abstract}

Connected and cooperative driving requires precise calibration of the roadside infrastructure for having a reliable perception system. \chg{To solve this requirement in an automated manner}, we present a \chg{robust} extrinsic calibration method for automated geo-referenced camera calibration. \chg{Our method requires a} calibration vehicle equipped with a combined GNSS/RTK receiver and an inertial measurement unit (IMU) for self-localization. \chg{In order to remove any requirements for the target's appearance and the local traffic conditions, we propose a novel approach using hypothesis filtering}. Our method does not require any human interaction with the information recorded by both the infrastructure and the vehicle. \chg{Furthermore, we do not limit road access for other road users during calibration}. We demonstrate the feasibility and accuracy of our approach by evaluating our approach on synthetic datasets as well as a real-world connected intersection, \chg{and deploying the calibration on real infrastructure}. Our source code is publicly available\footnote{\url{https://github.com/Tuxacker/infrastructure_calibration_toolbox}}.

\end{abstract}

%%%%%%%%%%%%%%%%%%%%%%%%%%%%%%%%%%%%%%%%%%%%%%%%%%%%%%%%%%%%%%%%%%%%%%%%%%%%%%%%
\section{Introduction}

The research on automated driving topics has recently expanded beyond the automated vehicle itself and started including other sources of environment perception, mainly stemming from connected infrastructure, e.g.,~roadside units (RSUs)~\cite{buchholz2022int}. Such units mostly incorporate monocular cameras, which provide a simple and cheap way~\cite{rinner2008introduction} of environment perception. The information provided by the RSUs can be integrated into downstream tasks like road user trajectory prediction and cooperative trajectory planning~\cite{mertens2022cooperative}. Therefore, the localization accuracy of road users needs to be sufficient to avoid dangerous situations due to inaccurate data, assuming a high calibration quality of the infrastructure. Furthermore, if the RSU infrastructure is going to be installed in a place with heavy traffic, creating a constrained environment for precise calibration may be challenging. As an example, classical calibration methods mostly require manually placed calibration targets in the sensor's field of view (FoV)~\cite{andrew2001multiple}. The placement of such targets in a heavily congested intersection may not be viable in every situation, making manual targets unsuitable for performing extrinsic calibration in the field.
\begin{figure}
    \centering
    \includegraphics[width=0.475\textwidth]{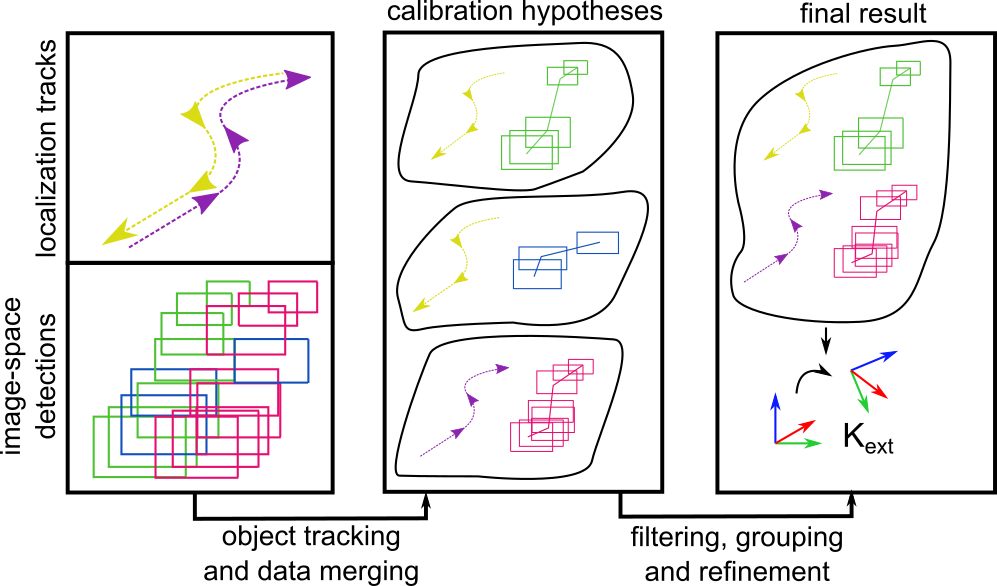}
    \caption{Approach overview. From left to right, tracked object bounding boxes extracted from camera images and vehicle localization data are recorded. Data from both domains are then merged based on the recorded timestamps to create extrinsic calibration hypotheses. The hypotheses are grouped and filtered to remove tracks of other road users. Finally, the best-performing hypothesis group is merged and refined to obtain the final calibration $\boldsymbol{K}_\text{ext}$.}
    \label{fig:overview}
\end{figure}

There is prior work on explicitly adapting the calibration for infrastructure, e.g.,~\cite{datondji2016rotation, zhu2016robust} to reduce human interaction and simplify the process. However, these methods require overlapping fields of view, implicating multiple sensors and thus driving up the infrastructure costs. Furthermore, in the case of temporary infrastructure, like at construction sites, only one RSU may be deployed, and a second sensor for calibration may not be available.%Another method for infrastructure calibration is described in~\cite{muller2019laci}, however it is limited to lidar sensors only.
~As an alternative, feature-based calibration approaches emerged in the literature. Edge-based approaches like \cite{dubska2014fully,dubska2014automatic} use object edge features to estimate an extrinsic calibration while under traffic by using the traffic itself as a calibration target. In contrast to other techniques, the resulting calibration is not geo-referenced, thus hindering the conversion of the detected traffic participants into the coordinate system of an automated vehicle.
%For moving sensor platforms \cite{wodtko2021globally} proposes an algorithm that uses the motion of the sensor platform for automated calibration and re-calibration. This class of algorithms can unfortunately only used for intelligent vehicles (IVs), as infrastructure sensors are inherently static.
~Finally, deep learning approaches like \cite{li2021deepi2p} have been proposed, where an extrinsic calibration is obtained from camera data and intelligent vehicle (IV) lidar sensor data by means of a neural network. \chg{With \cite{li2021deepi2p} reporting calibration errors of over $1\si{m}$, these approaches are unsuitable for cooperative driving scenarios. As an example, a vehicle detection with such localization errors may be erroneously assigned to an incorrect lane.}

We propose an automated single monocular camera calibration \chg{approach} designed for live traffic conditions, \chg{addressing the limitations of the previously mentioned methods}. Our method uses a GNSS/RTK- and IMU-equipped vehicle, which records its localization data while driving through the FoV of the sensor under calibration. %\chg{If the vehicle is driving autonomously, the entire calibration process is fully automatic}.
By combining the recorded vehicle localization data with synchronously recorded camera data, an extrinsic calibration can be estimated without any human interaction. Our approach consists of data pre-processing, hypothesis generation, filtering/grouping, and refinement steps. Fig.~\ref{fig:overview} gives a visual overview of the proposed algorithm. During the initial data pre-processing, objects in the camera images are detected with a neural object detection network. These objects are then tracked over time in pixel coordinates. In the hypothesis generation step, the tracks in pixel coordinates are combined with synchronized localization data from the calibration vehicle to create association hypotheses. By estimating the calibration from each hypothesis and subsequent filtering and grouping, the tracks most likely associated with the GNSS/RTK- and IMU-equipped vehicle are obtained, while tracks belonging to other traffic participants are removed. During the refinement step, the remaining hypotheses are combined, and the merged calibration from all hypotheses is refined.%This is performed by matching the projection of the 3D box, defined by the vehicle's dimensions, with the corresponding 2D bounding box obtained from the object detector.
~The refinement step yields the final extrinsic calibration. Our experiments using simulated data generated with DeepSIL~\cite{st2021sil} and CARLA~\cite{Dosovitskiy17} and real-world data from our test site~\cite{buchholz2022int} show sufficient accuracy for automated driving applications. % \chg{allowing us to deploy and use the obtained calibration results on our test site infrastructure}.
Our contributions are summarized as follows:
%\begin{itemize}
    %\item An automated approach for geo-referenced infrastructure calibration.
    %\item \chg{A novel hypothesis filtering scheme enabling the use of standard camera calibration algorithms like \cite{lepetit2009epnp} on unmodified and unfiltered data from the infrastructure.}
    %\item Only the sensor to be calibrated is needed, other infrastructure sensors are not required.
    %\item Minimal requirements for the calibration vehicle, as only a localization sensor (e.g.,~GNSS/RTK+IMU unit) is required, \chg{and no requirements for the appearance of the calibration vehicle.}
    %\item No requirements for the sensor environment, \chg{therefore} other road users may freely use the area captured by the sensor \chg{during the calibration process}.
    %\item Fast execution speed, as the calibration can be calculated within seconds after obtaining the calibration data .
%\end{itemize}
\begin{itemize}
    \item An automated approach for geo-referenced infrastructure calibration based on a novel hypothesis filtering scheme enabling the use of standard camera calibration algorithms like \cite{lepetit2009epnp} on unmodified and unfiltered data from the infrastructure.
    \item Minimal requirements for the calibration vehicle, as only a localization sensor is required and the vehicle size has to be known, and almost no requirements for the appearance of the calibration vehicle as well as the sensor environment, \chg{therefore} other road users may freely use the area captured by the sensor \chg{during the calibration process}.
    \item Experiments on synthetic and real-world data show applicability in automated and cooperative driving tasks.
\end{itemize}
%The source code of our approach will be made publicly available upon publication.

\section{Related Work}

The calibration of camera sensors is a widely explored topic, which can be further divided into four different types. Namely, the methods can be split into target-based like ours, feature-based, neural-network-based, and motion-based approaches. 

The most classic approaches are target-based, e.g.,~\cite{andrew2001multiple}, and use a chessboard pattern or other target types to estimate intrinsic and extrinsic calibration parameters. A more specialized approach targeting road infrastructure was presented in \cite{datondji2016rotation}, where a pair of fisheye cameras with large overlapping fields of view were calibrated using vehicle calibration targets. In a similar fashion, our approach does not need static targets that may hinder the traffic during calibration.

Feature-based approaches adopt various image features as 2D-3D-correspondences. The algorithm presented in \cite{ataer2014calibration} employs a mobile RGB-D camera setup to scan the environment and utilizes this data to create a 3D environment model of the observable area. Then, the 3D model features are matched with 2D feature descriptors to find the closest matching keyframe from the scan. Another method described in \cite{dubska2014fully} extracts and tracks object edges of moving vehicles. The edge features are then subsequently used to estimate the camera's vanishing points and calculate the extrinsic camera transformation matrix into a local, non-geo-referenced coordinate system. A newer idea proposed in \cite{bartl2021automatic} employs a neural network trained to output landmark positions of specific vehicle models. The spatial relations between the landmarks are then compared with distances obtained from measuring the landmark distances of an accurate 3D model of the selected vehicle type to find a suitable camera calibration, which, however, is not geo-referenced. Finally, the method described in \cite{ts2022semantic} applies semantic segmentation networks to segment camera images as well as lidar scans of the environment and then minimizes the visual error between the image segmentation and a rendered view of the segmented point cloud by adapting the initial camera transformation parameters. In contrast to the presented methods, our method is global, i.e.,~it does not require an initial calibration estimate to work while also providing a geo-referenced global position.

With the recent emergence of neural networks to perform various tasks, direct 2D-3D matching algorithms were proposed. The network presented in \cite{feng20192d3d} directly predicts the similarity between point cloud patches and image patches. \cite{yu2020monocular} find correspondences between 2D line features obtained from a neural network with 3D line features calculated from a lidar scan, while \cite{li2021deepi2p} utilize a classification network to classify points being inside or outside of the camera frustum and then solve an inverse camera projection problem. Our proposed method, in comparison, does not use neural networks to create 2D-3D correspondences but only to perform object detection, which results in smaller calibration errors.

Another distinct algorithm type present in literature is motion-based calibration~\cite{horn2021online,wodtko2021globally,wei2018calibration,wise2020certifiably}, which exploits the movement of the sensor itself and is particularly suitable for automated vehicle calibration or online re-calibration. For example, \cite{horn2021online} employ visual SLAM to estimate the sensor movement and then optimize the loop closure between different time steps using an efficient dual quaternion representation of the extrinsic rotation and translation, which allows for real-time re-calibration. This approach type is not applicable to infrastructure, as it is stationary and captures a static view of the environment.
%We build on previous methods by fully automating a target-based calibration procedure.

\section{Calibration Method}

In the following subsections, we formulate our calibration objective and describe every step of our algorithm. An overview of the algorithm is shown in Fig.~\ref{fig:overview}.

\subsection{Problem Formulation}

A standard perspective camera model~\cite{andrew2001multiple} is defined as follows:
\begin{equation}
\label{eqn:problem}
	{\begin{bmatrix}
	u^* \\ v^* \\ c
	\end{bmatrix}} =
	%\underbrace{
	\boldsymbol{K}_\text{int} \;
	\overbrace{
	\begin{bmatrix}
	\boldsymbol{R} & \boldsymbol{t} \\
	\boldsymbol{0} & 1
	\end{bmatrix}}^{:= \boldsymbol{K}_\text{ext}}
	%}_{:= \boldsymbol{P}}
	{\begin{bmatrix}
	x \\ y \\ z \\ 1
	\end{bmatrix}} \; ,
	{\begin{bmatrix}
	u \\ v \\ 1
	\end{bmatrix}} = \frac{1}{c}
	{\begin{bmatrix}
	u^* \\ v^* \\ c
	\end{bmatrix}}
\end{equation}
where $[u, v, 1]^T$ denotes a homogeneous camera image coordinate of an image $\mathcal{I}$, and $[x, v, z, 1]^T$ is a homogeneous point coordinate of a point in 3D world space. Our goal is to estimate $\boldsymbol{K}_\text{ext}$ containing a rotation $\boldsymbol{R}$ and translation $\boldsymbol{t}$, which can transform between world/UTM~\cite{grafarend1995optimal} and sensor coordinates. We assume the intrinsic camera matrix $\boldsymbol{K}_\text{int}$ is known. The data used for calibration consists of GNSS/RTK+IMU measurements in UTM coordinates $\mathbf{x}_t = [x_\text{UTM}\ y_\text{UTM}\ z_\text{UTM}\ \alpha\ \beta\ \gamma]$ at timestep $t$, where $\alpha,\ \beta,\ \text{and}\ \gamma$ are the roll, pitch, and yaw, respectively. Furthermore, we utilize the vehicle box detections of the camera under calibration, where each detection $\mathbf{b}_{i, t} = [u\ v\ w\ h]$ has a unique ID $i$ and contains the box center coordinates $u$ and $v$, as well as the width $w$ and height $h$.

\subsection{Data Acquisition and Pre-processing}

At the beginning of the calibration pipeline, data from the sensor under calibration as well as the location of the target vehicle has to be recorded. Our algorithm only imposes two constraints on this recording process. Firstly, we require time synchronicity between camera and localization measurements, as the subsequent 2D-3D correspondences are based on the association of camera and vehicle data by their timestamps. %\chg{This requirement is easily enforced, as connected infrastructure for assisted driving is required to provide timestamp data with the same time reference as the vehicles using the data. Furthermore, as the target vehicle is equipped with a GPS receiver, GPS time can be used as a time reference, if the infrastructure is also equipped with GPS}.
\chg{This requirement is enforced by the connected infrastructure, as it is also required for assisted driving}. Secondly, we require at least two different traversals of the sensor's FoV, such that the target vehicle is not visible or detectable in the camera image between traversals. This fact is used later to filter out false positive hypotheses in Sec.~\ref{sec:filtering}. After collecting the calibration data, the data has to be processed for the subsequent steps.
To process the vehicle localization data, either the mean position of the recording or an externally defined reference point is subtracted from each position measurement. This helps the numerical stability, as the recorded UTM coordinates may be in the range of $10^5-10^7$, which can lead to numerical instabilities during later matrix (inversion) operations. To process the image data, object-level features are extracted. The feature extraction is performed by a neural object detection network, which returns bounding boxes for all vehicles in an image. This leaves us with vehicle bounding boxes $\mathbf{b}_{i, t}$ of all vehicles that passed the sensor infrastructure during a single time step $t$ of the recording, with each one having a unique ID $i=1\hdots N$, including our calibration target vehicle.

\subsection{Object Tracking in Image Space}

Following the bounding box extraction, the object boxes are associated over all time steps, thus tracking the objects over the recording period. Usually, a multi-object tracker (MOT)~\cite{bar2004estimation} is used for such tasks, however, this requires the sensor measurements to be in a metric space in order to apply the underlying kinematic models, i.e.,~CV/CA/CTRA~\cite{cvmodel}. As in our case, the transformation to the metric space is not known yet; such trackers can not be adopted for our purpose. Instead, we rely on a model-less tracking approach. Therefore, we use the Hungarian Algorithm~\cite{crouse2016hung} to find the most cost-effective box-to-box association between two frames. In our case, we define our association cost as follows:
\begin{flalign}
    d(\mathbf{b}_1, \mathbf{b}_2) = 
    \begin{cases}
    1 - \text{DIoU}(\mathbf{b}_1, \mathbf{b}_2)&,\text{if}\, \text{IoU}(\mathbf{b}_1, \mathbf{b}_2) > \mathrlap{0}\\
    2&,\text{else}
    \end{cases}
    \label{eqn:assoc_cost}
\end{flalign}
where IoU is the intersection-over-union between two boxes and DIoU is the distance-IoU introduced in~\cite{zheng2020distance}, which additionally penalizes the distance between bounding box centers and can take values between $0$ and $2$. While using Eq.~\ref{eqn:assoc_cost}, we assume that the camera frame rate is high enough so that two consecutive bounding boxes of the same vehicle partially overlap if the vehicle obeys the speed limit, otherwise incorrect object associations may occur at a much higher rate. This implies that object boxes not overlapping with another box are newly detected objects. Furthermore, by using the DIoU, we additionally penalize associating object boxes with vastly different sizes and center locations. Using the Hungarian algorithm and preventing association between two objects if their cost equals 2, we can add new box measurements to existing tracks and set up new tracks if no matching box from the previous frame is found. If a track from the previous frame has no associated boxes in the current frame, it is linearly extrapolated. This extrapolation is performed until the track can be associated with a newly detected box or the amount of extrapolated timestamps exceeds a threshold, in which case the track is removed from the current list of tracked objects. By applying this tracking procedure to the data of the entire recorded sequence, it is possible to reconstruct the path of each vehicle that drove through the sensor's FoV. As the tracking is performed offline, we can remove all extrapolated box positions from the tracks. At the end of this step, we have a list of $M$ object tracks $\mathcal{T}=\{\mathbf{o}_1\,\hdots\,\mathbf{o}_M\}, \mathbf{o}_i=\{\mathbf{b}_{i,t}\}_{t=t_{\text{start},i}}^{t_{\text{end},i}}$ belonging to uninvolved traffic and the target vehicle.

\subsection{Hypothesis Creation}
\label{sec:hypot}
After obtaining all object tracks $\mathcal{T}$ from the images of the sensor under calibration, the calibration vehicle tracks need to be detected and separated from other object tracks. We formulate this task as a kind of hypothesis filtering, where each object track may hypothetically be the target vehicle. To construct a hypothesis $\mathbf{h}_i=\{\mathbf{x}_t, \mathbf{b}_{i, t}\}_{t=t_{\text{start},i}}^{t_{\text{end},i}}$, we associate a given object track $\mathbf{o}_i$, where we assume the object location to be the bounding box center, with the recorded GNSS/RTK+IMU data $\mathbf{x}$ from the same timestamps $t\in[t_{\text{start},i};t_{\text{end},i}]$, yielding 2D-3D correspondences between the image and world coordinate frames. The width and height contained in $\mathbf{o}_i$ are not considered for the initial estimate.

After creating a hypothesis out of every object track, initial filtering can be performed. During the initial filtering, all hypotheses which do not contain enough data points (the EPnP algorithm used requires at least 4 correspondences) or have short track lengths in either 2D or 3D coordinates (i.e.~the observed vehicle didn't move) are discarded. Then, all remaining hypotheses are accepted to obtain a first estimate of $\boldsymbol{K}_\text{ext}$ for each hypothesis using the EPnP algorithm~\cite{lepetit2009epnp}. As this approach is negatively affected by outliers in the input data, a RANSAC scheme~\cite{zuliani2009ransac} is employed to filter out such outliers. If the median reprojection errors of the estimates exceed a threshold, the hypothesis is discarded. To further narrow down the hypotheses, the camera position is obtained with $\boldsymbol{x}_\text{cam} = -\boldsymbol{R}^T\boldsymbol{t}$. The hypothesis is then also discarded if the camera position is over a threshold distance $d_\text{thr}$ away from the path driven by the target vehicle or if the camera height is below ground. $d_\text{thr}$ can be in the range from $10\si{m}$ for real data to $30\si{m}$ for synthetic data. The remaining $N_f$ hypotheses form the set $\mathcal{H}=\{\mathbf{h}_i\}^{N_f}_{i=1}$.

\subsection{Hypothesis Grouping and Filtering}
\label{sec:filtering}
After determining the set of filtered, valid hypotheses $\mathcal{H}$, a unique solution has to be found. At this stage, $\mathcal{H}$ should contain at least two or more tracks belonging to the target vehicle, depending on the number of passes along the infrastructure. However, incorrect hypotheses belonging to other vehicles may still be contained in the set. This can occur if the trajectory of the calibration vehicle driving outside the sensor's FoV closely resembles the trajectory of another vehicle passing the sensor, which is most common on straight stretches of road.% where every traffic participant drives on similar paths. 
~To remove incorrect hypotheses, clustering of similar hypotheses is performed in two parts. In the first part, the similarity between two hypotheses $\mathbf{h}_i$ and $\mathbf{h}_j$ is determined by the following scores:
\begin{itemize}
    \item Outlier ratio: The object track contained in  $\mathbf{h}_i$ is projected into world space with the calibration estimate $\boldsymbol{K}_{\text{ext},j}$ of $\mathbf{h}_j$. The ratio $r_\text{out}$ of points lying outside the camera frustum of hypothesis $\mathbf{h}_j$ is determined.
    \item Overlap ratio: The localization track contained in $\mathbf{h}_i$ is projected into image space with the calibration estimate $\boldsymbol{K}_{\text{ext},j}$. Then, the ratio $r_\text{ov}$ of 2D localization points within $k$ pixels of any bounding box in the calibration sequence is calculated. 
    \item Rotational similarity: We define and obtain the rotational similarity $r_\text{sim}=tr(\boldsymbol{R}^{-1}_j\boldsymbol{R}_i) / 3$, where $\boldsymbol{R}_{\{i,j\}}$ is the upper left matrix of $\boldsymbol{K}_{\text{ext},{\{i,j\}}}$, as defined in Eq.~\ref{eqn:problem}.
\end{itemize}
If all three scores lie within specified ranges described by a set of hyper-parameters, the estimated $\boldsymbol{K}_\text{ext}$ of two hypotheses and, therefore, the hypotheses themselves are considered similar. By comparing every pairing possible in $\mathcal{H}$, we can build a hypothesis graph where similar hypotheses are connected by edges. In the second step, all unconnected nodes are discarded, as we require at least two traversals of the camera environment in order to create a groupable sub-graph in the first step. The remaining sub-graphs are fed to the DBSCAN~\cite{schubert2017dbscan} clustering algorithm to group all hypotheses with similar camera coordinates, with most to all outliers removed. This results in a list of different hypothesis groups, where each group contains only similar hypotheses. If no hypotheses are left at this stage, it indicates an insufficient count of calibration vehicle traversals.

\subsection{Result Refinement}
\label{sec:refinement}
%\begin{figure}
%    \centering
%    \includegraphics[width=0.45\textwidth]{export1.png}
%    \caption{Overview of the refinement procedure in image space (left) and bird's eye view (right). A new correspondence is found by finding the vehicle corner point closest to the bottom box edge (denoted in cyan). The associated bounding box point is determined by backprojecting the corner point into the image and finding its closest point on the bottom box edge (denoted in light green). The length of the dark green line represents the localization error made by using a specific calibration. This error is denoted as $\delta_p$ in Sec.~\ref{sec:eval}. The vehicle image was generated with CARLA~\cite{Dosovitskiy17}.}
%    \label{fig:refinement}
%\end{figure}
\begin{figure*}
    \centering
    \includegraphics[width=0.75\textwidth]{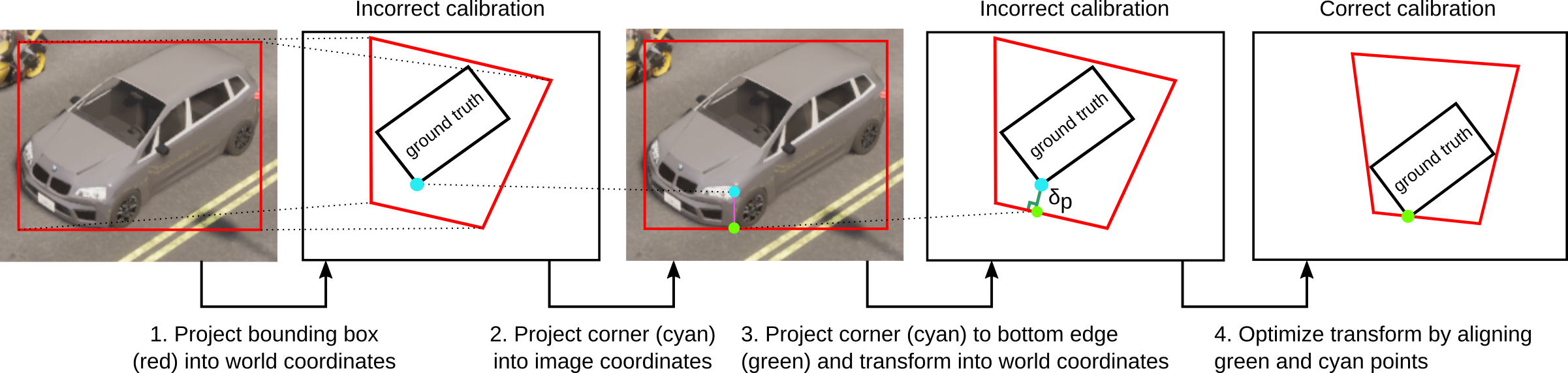}
    \caption{Overview of the refinement procedure in image space and bird's eye view. A new correspondence is found by finding the vehicle corner point closest to the bottom box edge (denoted in cyan). The associated bounding box point is determined by backprojecting the corner point into the image and finding its closest point on the bottom box edge (denoted in light green). The length of the dark green line represents the localization error made by using a specific calibration. This error is denoted as $\delta_p$ in Sec.~\ref{sec:eval}. The vehicle image was generated with CARLA~\cite{Dosovitskiy17}.}
    \label{fig:refinement}
\end{figure*}
After obtaining valid hypothesis groups, the data inside these groups can be refined and merged. Using merged tracks containing different trajectories may help minimize the heteroscedastic errors introduced by missing depth information.% and an insufficient amount of data along a specific dimension. An example of such an axis is the ``depth'' axis of the camera.
~Furthermore, in Sec.~\ref{sec:hypot}, the center of the bounding box and the geometric center of the vehicle are used as correspondences. However, this may not necessarily be accurate.

We, therefore, propose a refinement step that exploits knowledge about the size of the target vehicle to reduce the error caused by the previous assumption. By using the orientation information contained in $\boldsymbol{x}_t$ and the known vehicle dimensions, we can calculate the footprint of the vehicle, i.e.,~the world coordinates of its corners in the ground plane. Due to geometrical constraints, one of these ground points must lie on the bounding box's bottom edge.%This point can be used to create new, improved correspondences.
~For refinement, the bounding box of the vehicle is projected onto the ground plane resulting in a quadrilateral. The projection is made with one of the previously obtained hypothetical transformations. The required ground plane is estimated by performing a principal component analysis (PCA)~\cite{abdi2010principal} of the recorded localization track. The vehicle corner point with the lowest distance to the projected bottom line of the bounding box rectangle is chosen as the new vehicle reference point $\mathbf{x}_{i,t,n}$ (cyan point in Fig.~\ref{fig:refinement}). To get the corresponding bounding box point $\mathbf{b}_{i,t,n}=[u_n\ v_n]$ (light green point in Fig.~\ref{fig:refinement}), $\mathbf{x}_{i,t,n}$ is projected back into the 2D bounding box, where its closest point on the bottom box edge is found. This process is visualized in Fig.~\ref{fig:refinement}. Alternatively, the optimization approach described in~\cite{masi2021augmented} can be used to obtain refined point pairs. After obtaining the refined point pairs $\{\mathbf{x}_{i, t,n},\,\mathbf{b}_{i, t,n}\}$, the line-point registration described in \cite{Briales_2017_CVPR} is applied to calculate an optimized transformation which forces the physical boundaries of the vehicle to align with the bounding box bottom line projected onto the ground plane, improving the estimated calibration. By performing this refinement on each hypothesis separately, then merging the data of all hypotheses tracks of a group and repeating refinement on the ensemble hypothesis, the extrinsic calibration $\boldsymbol{K}_\text{ext}$ is obtained for each group. By measuring the distance between the point correspondences used for refinement after performing the optimization step, the transformation with the smallest average distance between the projected bounding box point and vehicle corner point $\delta_p$, denoted by the dark green line in Fig.~\ref{fig:refinement}, is chosen as the final $\boldsymbol{K}_\text{ext}$. The merged calibration may be discarded if the average error $\delta_p$ of one specific track is lower than $\delta_p$ of the merged calibration when evaluating on the merged tracks.

\section{Evaluation}

In order to test our approach, a number of different evaluation protocols were applied. The evaluation was performed with both real-world sequences from our test site as well as synthetic data. In the following, we provide a detailed description of the evaluation setups and present the results.

\subsection{Data sources and Evaluation Setup}

%We aim to assess both the theoretical performance of our approach as well as its applicability to real-world deployments,a hybrid evaluation setup was chosen.

We evaluate our approach on both real and synthetic data. First, we employ an extended version of the DeepSIL~\cite{st2021sil} and CARLA~\cite{Dosovitskiy17} simulators to generate synthetic localization and detection data as input for the algorithm. Second, we rely on real-world sequences from our test site~\cite{buchholz2022int}. \chg{The calibration obtained from the real-world sequences is deployed at the test site and is utilized for cooperative driving tests.}

The DeepSIL simulator uses the digital map of our real-world test site~\cite{buchholz2022int} to create a digital twin and simulate scenarios with different virtual vehicles which pass it. During the simulation, one actor vehicle is chosen as the calibration vehicle, and its world position during every time step is recorded. To generate image-space detections, three virtual cameras are defined in the digital map. Using the defined camera transforms, detection bounding boxes for each vehicle inside the virtual sensor's FoV are generated. As real-world detections and measurements are not exact and always come attached with some type of detection probability and measurement noise, a simple, normally distributed noise is added to the generated data. We added Gaussian noise with $\sigma=0.075\si{m}$ to both the x- and y-position of the simulated calibration vehicle, therefore emulating noisy measurements of the localization track. Noisy bounding box detections can also be generated if the additive noise significantly exceeds the measurement noise of a real-world localization unit. For each type of simulated measurement, its corresponding generation timestamp is recorded as well, as it is required for hypothesis creation. The tracking, hypothesis creation, filtering, and refinement steps are performed directly on the resulting synthetic data, which is used as an input to the algorithm.

The CARLA configuration is similar to the DeepSIL setup. However, the map does not correspond to a real-world location. Furthermore, we do not add any noise to the data generated by CARLA to evaluate our approach's performance in a controlled environment and simulate ideal measurements while having access to different scenarios involving signalized intersections and vulnerable road users.

We also recorded multiple calibration sequences of up to 15 minutes in length on a real intersection located in Ulm-Lehr, Germany. The intersection consists of multiple cameras with a resolution of 1920x1200 and a frame rate of 10 fps, where some cameras have partially overlapping fields of view, are rotated into a portrait orientation, and have different focal lengths. Therefore wide-angle, standard, and telephoto perspectives are available and evaluated in Table~\ref{tbl:real}.% and shown in Fig.~\ref{fig:cameras}.
~A test vehicle equipped with a GNSS/RTK+IMU localization system is used as a target vehicle providing high-precision localization data with a rate of $50\si{Hz}$ \chg{and localization errors below $5\si{cm}$}.
The synchronization between both data sources is ensured by using GPS time as the reference time, \chg{as both our vehicle and test site are equipped with GPS receivers providing the same time reference, with measured clock offsets below $2\si{ms}$, enabling the execution of cooperative maneuvers}. Evaluation with real-world sequences is performed \chg{offline after recording in} the same way as with synthetic data, with the only difference being the additional feature extraction step, in which objects are detected in the camera images to obtain object bounding boxes. For this purpose, we rely on YOLOv5s network~\cite{glenn_jocher_2022} trained on the BDD100K dataset~\cite{yu2020bdd100k}. %As no real ground truth calibration was available for these cameras, the resulting calibrations were compared to manual calibrations obtained from manually annotating known landmarks of the calibration vehicle in the recorded video sequences.

\subsection{Results discussion for synthetic and real calibration sequences}
\label{sec:eval}
Table~\ref{tbl:synthetic} summarizes the evaluation results with synthetic data generated by DeepSIL and CARLA.
Due to the use of bounding box object data, the object's outline is unavailable. Furthermore, the box corners usually do not align with object corners, making them unsuitable for distance error calculation. In order to circumvent this lack of information, we apply the approach from Sec.~\ref{sec:refinement} shown in Fig.~\ref{fig:refinement}. Here, we project the bottom box edge onto the ground plane and measure the minimal distance $\delta_p$ to the closest ground point of the calibration vehicle, which is obtained from the localization measurements.
We provide the mean and maximal absolute distance errors $\delta_{p,m}$ and $\delta_{p,w}$ as we want to assess the usability of estimated calibrations for real cooperative driving use cases and deployments instead of just comparing the camera position to a ground truth position. We also report the mean and maximum relative distance errors $e_m$ and $e_w$, which are obtained by dividing $\delta_{p,\{m,w\}}$ by the distance between the vehicle corner point and camera.
% similar to \cite{dubska2014fully}.
The evaluation is always performed on the entire merged data of all calibration vehicle tracks.

\begin{table}[h]
    \caption{Synthetic Evaluation results for DeepSIL and CARLA.}
    \label{tbl:synthetic}
    \centering
    %. We report the average translation error in \si{\cm} and rotation error in degrees of each scene. The 10 results with the lowest final loss over a total of 30 runs were used in order to find the parameter sets that mostly likely converged to the global minimum. For KITTI, both the semi-automatic as well as the fully automated pipelines were tested.}
    \begin{tabular}{lcccc}
        \toprule
        \textbf{Scene} & SIL-North & SIL-South & SIL-West & CARLA\\
        \midrule
        $e_m$ (\%) & 0.83 & 1.39 & 0.22 & 0.17\\
        \midrule
        $e_w$ (\%) & 2.80 & 2.22 & 1.80 & 0.68\\
        \midrule
        $\delta_{p,m}$ (\si{m}) & 0.36 & 0.54 & 0.06 & 0.087\\
        \midrule
        $\delta_{p,w}$ (\si{m}) & 1.52 & 0.86 & 0.4 & 0.237\\
        \bottomrule
    \end{tabular}
\end{table}
From the results, we can clearly see the performance of our approach. The results are promising when looking at the CARLA results obtained from ideal, noise-free data. However, an average $\delta_{p,m}$ of around 0.09\si{m} suggests a small systematic error. This error stems from the refinement step, where the distance between the projected bottom edge and the closest ground point is minimized. An error is introduced as this is done by linearly projecting the ground point onto the bottom edge from the currently estimated perspective instead of the ground truth perspective. The DeepSIL results show higher errors than the CARLA results and similar values to Table~\ref{tbl:real}, which can be explained by the added noise. %which is inherently present in real-world recordings.
~The comparatively higher mean error for the virtual ``South'' camera stems from the fact that this camera is placed almost away 30 meters from any road segment. Therefore, measurements close to the camera are not present, impacting the calibration quality. 

%As the ground truth calibration for the test site is not known, the average deviation from a reference calibration has been provided. The reference calibration was obtained by manually annotating landmarks on the calibration vehicle. In addition to the synthetic data, the calibration results from the real test site recordings is displayed in Table~\ref{tbl:real}, where the focal length $f$ of each camera is reported in addition to the metrics used in Table~\ref{tbl:synthetic}. 
\begin{table}[h]
    \caption{Evaluation results from real-world sequences.}
    \label{tbl:real}
    \centering
    %. We report the average translation error in \si{\cm} and rotation error in degrees of each scene. The 10 results with the lowest final loss over a total of 30 runs were used in order to find the parameter sets that mostly likely converged to the global minimum. For KITTI, both the semi-automatic as well as the fully automated pipelines were tested.}
    \begin{tabular}{lccccc}
        % Wide: SPU21, 2nd recording 040822
        % Tele1; SPU01, 2nd recording 040822
        % Tele2; SPU22, 2nd recording 040822
        % Portrait; SPU31, 2nd recording 040822
        % Regular; SPU62, 2nd recording 040822
        \toprule
        \textbf{Camera name} & Wide & Tele 1 & Tele 2 & Portrait & Regular\\
        \midrule
        $f$ & 815.5 & 2174 & 2186 & 1067 & 1422 \\
        \midrule
        $e_m$ (\%) & 0.91 & 0.67 & 0.29 & 0.55 & 0.65\\
        \midrule
        $e_w$ (\%) & 2.94 & 2.42 & 2.33 & 4.78 & 3.17\\
        \midrule
        $\delta_{p,m}$ (\si{m}) & 0.26 & 0.37 & 0.14 & 0.24 & 0.32\\
        \midrule
        $\delta_{p,w}$ (\si{m}) & 0.50 & 1.49 & 1.03 & 2.18 & 1.44\\
        %\midrule
        %$\delta_{p,ref,m}$ (\si{m}) & 0.09 & 0.03 & 0.001 & 0.07 & 0.22\\
        \bottomrule
    \end{tabular}
\end{table}

\begin{figure}
    \centering
    \includegraphics[width=0.325\textwidth]{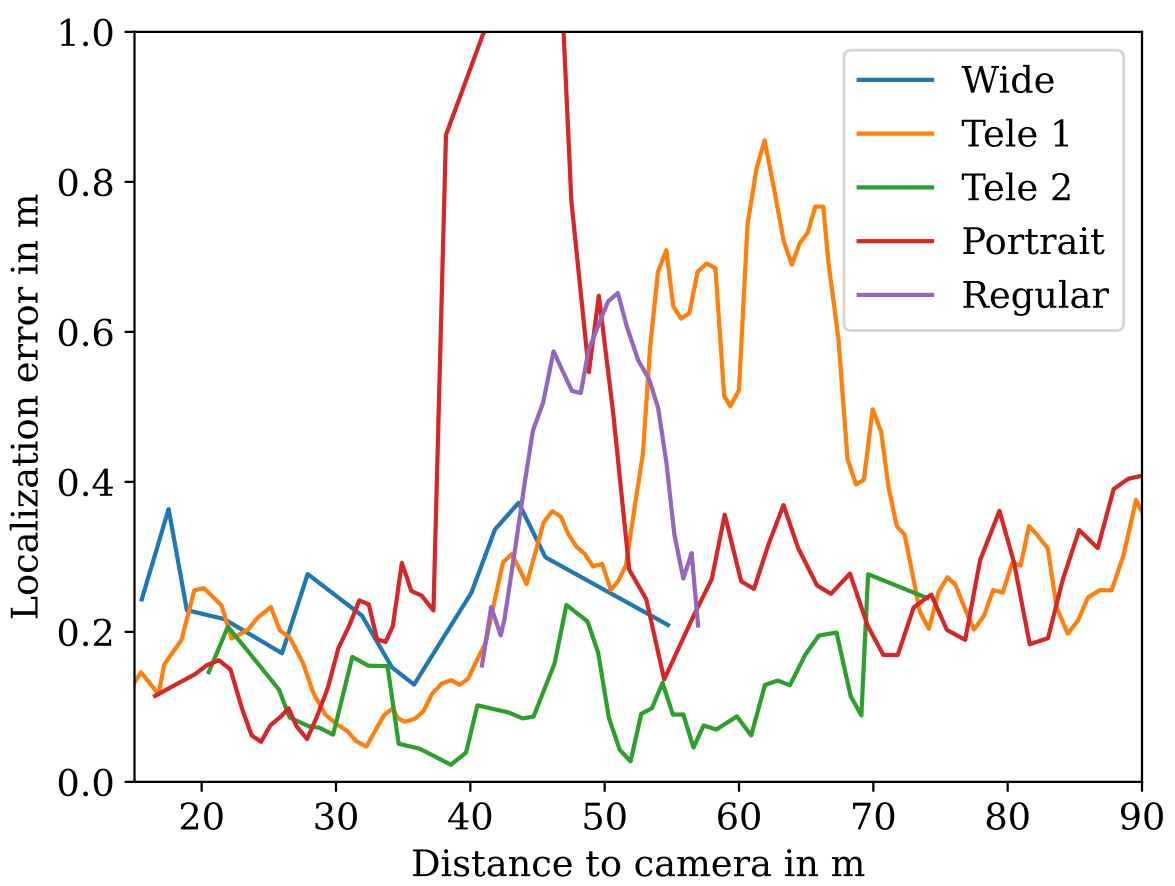}
    \caption{Calibration vehicle localization error $\delta_p$ mean values over 0.5\si{m} distance intervals plotted over the distance between camera and vehicle. For the ``Portrait'' camera, the maximal error of $2.0 \si{m}$ is reached at $45 \si{m}$ and therefore clipped for better visualization.}
    \label{fig:errors}
\end{figure}

For the real-world sequences, in addition to the metrics reported in Table~\ref{tbl:synthetic}, the previously measured focal length $f$ contained in $\boldsymbol{K}_\text{int}$ is reported in order to differentiate the different cameras as well as demonstrate the applicability of our approach to different camera configurations. 
%Considering the results in Table~\ref{tbl:real}, our approach also works in a real use case.
When comparing the results of the different cameras, we can see that each camera, independent of its focal length, achieves mean distance errors below 0.4\si{m}, which is acceptable for automated scenarios, as the planning module of an automated vehicle can be adjusted to add 0.4\si{m} safety margins. Furthermore, the error does not increase significantly with greater distances between object and camera, as shown in Fig.~\ref{fig:errors}.%The maximal errors mostly occur at camera distances greater than 40m, where small errors in the bounding box predictions lead to large localization errors due to the heteroscedasticity of the 2D-3D mapping.
~The high maximal error of the camera in portrait orientation is caused by the occlusion of vehicles due to trees and lamp poles in the scene, as described in Sec.\ref{sec:ablation}.
%Comparing the reference calibration to our estimated calibration, we can see that our automatic approach is matching the quality of a manual calibration, implying that no accuracy is lost when automating the calibration process.
When looking at automatic calibration approaches like \cite{dubska2014fully} and \cite{dubska2014automatic}, \chg{we see similar or better numbers in the used evaluation metrics. However, our approach can not be compared directly with previous works, as other automatic approaches do not provide geo-referenced calibrations and are not evaluated on the same dataset.}% we match previous results or even show a substantial improvement \chg{(g5 has a similar focal length to our Tele 2 camera, but reports $e_m=0.7\%$ and $e_w=4.0\%$)}, however, it should be noted that the evaluation is not performed on the same data and thus should be considered a qualitative comparison. \chg{A direct performance comparison with other approaches using common calibration benchmarks is not possible !REF, as the benchmarks do not contain the additional localization data and timestamps required by our method.}

%\begin{figure}
%    \centering
%    \includegraphics[width=0.4\textwidth]{camera_overview_iv.png}
%    \caption{Camera names, from left to right and top to bottom: Tele 1, Tele 2, Portrait, Wide, Regular. In view of Tele 1, a tree on the right side is covering the right lane at a distance, while the intersection view of Portrait is occluded by a tree and pole in the intersection area.}
%    \label{fig:cameras}
%\end{figure}

\subsection{Further Experiments}
\label{sec:ablation}

We repeated the DeepSIL simulation with different magnitudes for the position noise to evaluate our results beyond our provided localization error measurements. As seen in Table~\ref{tbl:noise}, the average and maximal relative errors $e_m$ and $e_w$ increase, but only up to a certain threshold, as the results for $\sigma_{pos}=0.2\si{m}$ and $\sigma_{pos}=0.5\si{m}$ are very similar. We explain this due to our application of a RANSAC-based transform estimation, which directly discards incorrect data points that cause excessive reprojection errors, making our algorithm more robust against low-quality input data. It should be noted, however, that in a real use case, the measurement errors of the localization unit are not randomly distributed but rather slowly drift over time, resulting in a small position bias, which has a smaller impact on the estimated transform.
\begin{table}[h]
    \caption{DeepSIL simulation with different additive noise for SIL-South scenario}
    \label{tbl:noise}
    \centering
    %. We report the average translation error in \si{\cm} and rotation error in degrees of each scene. The 10 results with the lowest final loss over a total of 30 runs were used in order to find the parameter sets that mostly likely converged to the global minimum. For KITTI, both the semi-automatic as well as the fully automated pipelines were tested.}
    \begin{tabular}{lccc}
        \toprule
         $\sigma_{pos}$ & 0.075 \si{m} & 0.2 \si{m} & 0.5 \si{m}\\
        \midrule
        $e_m$ (\%) & 1.39 & 2.22 & 2.30\\
        \midrule
        $e_w$ (\%) & 2.22 & 11.32 & 12.36\\
        %\midrule
        %$\delta_{p_m}$ (\si{m}) & 0.54 & 1.05 & 1.05\\
        \bottomrule
    \end{tabular}
\end{table}

In addition to the measurement noise experiment, we visualize the localization error in relation to camera distance for all real-world sequences in Fig.\ref{fig:errors}. It is visible that the error is not increasing significantly with increased distance to the camera, indicating that the rotation is estimated correctly. Furthermore, some regions of increased errors can be identified. In our case, the ``Portrait'' camera observes a light pole obstacle at a distance of approx.~45\si{m}, causing partial or missing detections of vehicles behind the obstacle, which results in high localization errors. The same applies to camera ``Tele 1'', where one lane is partially occluded by a tree at a distance of 60 \si{m}, while the error of camera ``Regular'' is caused by an uneven slope of the ground in front of the camera. %These obstacles can also be seen in Fig.~\ref{fig:cameras}.

\section{CONCLUSION}

We presented a new automated approach for geo-referenced infrastructure calibration based on hypothesis filtering. Our method only imposes minimal requirements on the calibration vehicles, is insensitive to other traffic participants, and is not dependent on human interaction. The calibration is performed by recording camera data and the position and orientation of the target vehicle while driving at least twice through the camera's FoV under calibration. In a pre-processing step, object tracks are extracted from the camera data. Then, the camera and vehicle tracks are combined during hypothesis generation. Subsequent filtering removes incorrect hypotheses, and in the final refinement step, the calibration is refined to take the 3D bounding box of the vehicle into account. An evaluation on synthetic and real-world data shows the effectiveness of our approach.%Future work could focus on further enhancements of the hypothesis filtering step and earlier inclusion of the full bounding box information during hypothesis creation.

\addtolength{\textheight}{-10cm}   % This command serves to balance the column lengths
                                  % on the last page of the document manually. It shortens
                                  % the textheight of the last page by a suitable amount.
                                  % This command does not take effect until the next page
                                  % so it should come on the page before the last. Make
                                  % sure that you do not shorten the textheight too much.

%%%%%%%%%%%%%%%%%%%%%%%%%%%%%%%%%%%%%%%%%%%%%%%%%%%%%%%%%%%%%%%%%%%%%%%%%%%%%%%%

%%%%%%%%%%%%%%%%%%%%%%%%%%%%%%%%%%%%%%%%%%%%%%%%%%%%%%%%%%%%%%%%%%%%%%%%%%%%%%%%

%%%%%%%%%%%%%%%%%%%%%%%%%%%%%%%%%%%%%%%%%%%%%%%%%%%%%%%%%%%%%%%%%%%%%%%%%%%%%%%%
%section*{APPENDIX}

%Appendixes should appear before the acknowledgment.

%%%%%%%%%%%%%%%%%%%%%%%%%%%%%%%%%%%%%%%%%%%%%%%%%%%%%%%%%%%%%%%%%%%%%%%%%%%%%%%%

\bibliographystyle{ieeetran}
\bibliography{bibliography}

\end{document}